\documentclass[11pt,a4paper]{article}
\usepackage[hyperref]{acl2020}
\usepackage{times}
\usepackage{latexsym}

\aclfinalcopy 
\def\aclpaperid{135} 

\usepackage{mathtools}
\usepackage{proof, lscape}

\usepackage{tikz}
\usetikzlibrary{shapes.geometric, positioning, shapes, fit, arrows, backgrounds}
\usetikzlibrary{intersections, calc}
\usepackage{gb4e}
\usepackage{multirow}

\usepackage{breakurl}
\hypersetup{breaklinks=true}

\newcommand{\rulelabelsize}{\scriptsize}
\newcommand{\FA}{\mbox{\rulelabelsize $>$}}
\newcommand{\BA}{\mbox{\rulelabelsize $<$}}
\newcommand{\SR}[1]{\begin{tabular}{c}$#1$\end{tabular}}

\newcommand{\comp}{\textsc{Comp}}
\newcommand{\bhline}[1]{\noalign{\hrule height #1}}  

\newcommand{\LF}[1]{\ensuremath{\mathbf{#1}}}
\newcommand{\fun}[2]{\ensuremath{\mathbf{#1}({#2})}}
\newcommand{\funt}[3]{\ensuremath{\mathbf{#1}({#2},{#3})}}
\newcommand{\funtheta}[2]{\ensuremath{\theta_{\text{#1}}(\text{#2})}}
\newcommand{\funtall}[2]{\ensuremath{\funt{tall}{#1}{#2}}}
\newcommand{\funadjer}[3]{\ensuremath{\exists \delta \,(\funt{#1}{#2}{\delta} \wedge \neg \,\funt{#1}{#3}{\delta})}}
\newcommand{\funtaller}[2]{\ensuremath{\funadjer{tall}{#1}{#2}}}
\newcommand{\funasadj}[3]{\ensuremath{\forall \delta(\funt{#1}{#2}{\delta} \rightarrow \funt{#1}{#3}{\delta})}}

\newcommand{\LabelYes}{\textit{yes}}
\newcommand{\LabelUnk}{\textit{unknown}}
\newcommand{\LabelNo}{\textit{no}}

\title{Logical Inferences with Comparatives and Generalized Quantifiers}

\author{Izumi Haruta$^{1}$ \\
        \texttt{haruta.izumi@is.ocha.ac.jp}
        \AND
        Koji Mineshima$^{2}$\\ 
        \texttt{minesima@abelard.flet.keio.ac.jp}
        \And
        Daisuke Bekki$^{1}$ \\ 
        \texttt{bekki@is.ocha.ac.jp}
        \AND
        $^{1}$\mbox{\rm Ochanomizu University, Tokyo, Japan} \\
        $^{2}$\mbox{\rm Keio University, Tokyo, Japan}
        }

\date{}

\begin{document}
\maketitle
\begin{abstract}
Comparative constructions
pose
a challenge in Natural Language Inference (NLI), which is the task of determining whether a text entails a hypothesis.
Comparatives are structurally complex in that they interact with other linguistic phenomena such as quantifiers, numerals, and lexical antonyms.
In 
formal semantics, there is a rich body of work on
comparatives and gradable expressions using the notion of degree. 
However, a logical inference system for comparatives
has not been sufficiently developed for use in the NLI task.
In this paper, we present a compositional semantics that maps various comparative constructions in English to semantic representations via
Combinatory Categorial Grammar (CCG) parsers and combine it with an inference system 
based on automated theorem proving.
We evaluate our system on three NLI datasets that contain complex logical inferences with comparatives, generalized quantifiers, and numerals. We show that the system outperforms previous logic-based systems as well as recent deep learning-based models.
\end{abstract}

\section{Introduction}

Natural Language Inference (NLI), or Recognizing Textual Entailment (RTE), is the task of determining whether a text entails a hypothesis and has been actively studied as one of the crucial tasks in natural language understanding.
In recent years, systems based on deep learning (DL) have been developed by crowdsourcing large datasets
such as Stanford Natural Language Inference
(SNLI)~\citep{bowman-etal-2015-large} and Multi-Genre Natural Language Inference (MultiNLI)~\citep{N18-1101}
and have achieved high accuracy.
NLI datasets focusing on complex linguistic phenomena, such as negation, antonyms, and numerals, have also been developed~\citep{naik-etal-2018-stress}.

However, it has been pointed out that
these datasets contain various biases that can be exploited by DL models~\citep{dasgupta2018evaluating,mccoy-etal-2019-right}, 
including easily classifying
numerical expressions in inference~\citep{liu-etal-2019-inoculation} 
and
answering by
only looking at a hypothesis~\citep{gururangan-etal-2018-annotation}.
This suggests that the success of
NLI models to date has been overestimated and that tasks remain unresolved. 

To handle inferences involving various linguistic phenomena,
there are also studies to probe the effects of additional training using artificially constructed data~\citep{dasgupta2018evaluating,richardson2020probing}. 
However, in the case of structurally complex inferences involving comparisons and numerical expressions,
there is a myriad of ways to combine possible inference patterns.
For example, consider the following inference.
\begin{exe}
    \ex \label{tab:map1}
    \scalebox{0.9}{
        {\renewcommand\arraystretch{1.3}
            \raisebox{-0.6cm}{
            	\begin{tabular}{cl}
    	            $P_1$: & John is taller than 6 feet.\\
                    $P_2$: & Bob is shorter than 5 feet.\\
  	            \hline
    	            $H$: & Bob is not taller than John. \ (Yes)\\
  	            \end{tabular}
  	        }
  	    }
  	}
\end{exe}

To correctly derive $H$ from $P_1$ and $P_2$, it is necessary to capture the predicate-argument structures of the sentences, antonyms (\textit{tall}, \textit{short}), numerical expressions, and negation.
Note that if the hypothesis sentence $H$ is changed to \textit{John is not taller than Bob}, the correct answer is not an entailment (Yes) but rather a contradiction (No);
even if numerical expressions are excluded, the number of combinations of sentence patterns that produces this kind of reasonable inference is enormous.

\begin{figure*}[t]
\scalebox{0.77}{
\hspace{0.3cm}
\begin{tikzpicture}
    \centering
    \tikzset{block/.style={rectangle, draw=black, line width=1pt, text width=1.7cm, text centered, rounded corners, minimum height=1.5cm}};
    \tikzset{block2/.style={rectangle, draw=black, fill=yellow!25, line width=1pt, text centered, rounded corners, minimum height=1cm}};
    \tikzset{plate/.style={rectangle, text centered, fill=white}};
    \tikzset{call/.style={ellipse callout, draw, line width=1pt, fill=cyan!25}}

    \tikzset{module/.style={rectangle, rounded corners=3mm, draw, fill=cyan!15}}

    \tikzset{DOT/.style={color=gray, very thick, dotted}}

    \tikzset{VECTOR/.style={color=black, very thick,->,>=stealth}}

    \node[block, text width=3.3cm, minimum height=2cm] (sen) {
        \centering
	    \scalebox{0.9}{
    	    {\renewcommand\arraystretch{1}
  			    \begin{tabular}{cc}
    				$P_1$: & John is tall.\\
        			$P_2$: & :\\
					\hline
    				$H$: & :\\
  				\end{tabular}
  			}
		}
    };
    \node[plate, text width=1.8cm, minimum height=0.5cm, above=1cm of sen.center, anchor=center] (W1) {\textbf{Sentences}};

    \node [module, below=0.8cm of sen] (A1) {
    \begin{tabular}{c}\textbf{Syntactic Parsing} \\ CCG parsers\end{tabular}};
    \draw[dashed, thick] (A1.west) -- (A1.east);    

    \node[block, text width=5.8cm, minimum height=2cm] (ccg) at (5.5,-2.8) {
		\centering
		\scalebox{0.72}{
        \infer[\BA]{S_{dcl}}{
        \infer[\LF{lex}]{NP}{
        \infer{N}{\mbox{John}}} & 
        \infer[\FA]{S_{dcl}\backslash NP}{
        \infer{(S_{dcl}\backslash NP)/(S_{adj}\backslash NP)}{\mbox{is}} & 
        \infer{S_{adj}\backslash NP}{\mbox{tall}}}}
		}
    };

  \node (SenToCCG) at (2.5,-1.3) {\textbullet};
  \draw[DOT] (A1.north east) edge (SenToCCG.center);

    \node[plate, text width=4.5cm, minimum height=0.5cm, above=1cm of ccg.center, anchor=center] (W2) {\textbf{(a)~CCG Derivation Trees}};
    \draw[VECTOR] (sen) -- ($(sen)!3.7cm!(ccg)$);
    \coordinate (P2) at (5.7,1);

    \node [module, right=0.5cm of sen] (A2) {
    \begin{tabular}{c}\textbf{Modifying Trees} \\ Tsurgeon\end{tabular}};
    \draw[dashed, thick] (A2.west) -- (A2.east);    

    \node (treemod) at (6.55,-0.3) {\textbullet};
    \draw[DOT] (A2.east) edge (treemod.center);

	\node[block2, text width=3cm, right=0.1cm of P2] (tree) {
    \textbf{Transformed CCG Trees}\\
	};
	\draw[VECTOR] (W2) -- (tree);
	
    \node[block, text width=5.7cm, minimum height=1.8cm, right=1.1cm of tree] (log) {
    	\centering
		\scalebox{0.85}{
    	    {\renewcommand\arraystretch{1}
  			    \begin{tabular}{cc}
    			    $P_1$: & $\exists \delta(\funtall{\LF{j}}{\delta} \wedge (\delta > \funtheta{tall}{U}))$\\
        			$P_2$: & :\\
					\hline
    				$H$: & :\\
  				\end{tabular}
  			}
		}
    };
    \node[plate, text width=5cm, minimum height=0.5cm, above=0.9cm of log.center, anchor=center] (W3) {\textbf{(b)~Logical Forms (A-not-A)}};
    \node (P3) at (9.6,1.0) {\textbullet};

    \node [module, below=1.0cm of P3] (A3) {
    \begin{tabular}{c}\textbf{Semantic Parsing} \\ ccg2lambda\end{tabular}};
    \draw[dashed, thick] (A3.west) -- (A3.east);    
    \draw[DOT] (A3.north) edge (P3.center);

    \draw[VECTOR] (tree) -- (log);

    \node[block2, text width=2cm, below=3.7cm of W3] (tptp) {
	\textbf{TPTP format}
	};

	\node[block2, left=0.8cm of tptp] (comp) {
	\begin{tabular}{c}\textbf{Axioms}\\ \comp \end{tabular}
	};

	\draw[VECTOR] (log) -- (tptp);
	\draw[VECTOR] (comp) -- (tptp);
	\node[block2, text width=2cm, right=1cm of tptp] (result) {
    \textbf{Yes, No,\\Unknown}
	};
	\draw[VECTOR] (tptp) -- (result);

    \node (P4) at (14.8,-2.63) {\textbullet};

    \node [module, right=2.4cm of A3] (Vampire) {
    \begin{tabular}{c}\textbf{Theorem Proving} \\ Vampire\end{tabular}};
    \draw[dashed, thick] (Vampire.west) -- (Vampire.east);    
    \draw[DOT] (Vampire.south) edge (P4.center);

\end{tikzpicture}
}
\caption{Overview of the proposed method.
The premises and hypothesis
are mapped to 
logical forms
based on A-not-A analysis
via CCG parsing and tree transformation;
then a theorem prover judges \LabelYes, \LabelNo, or \LabelUnk
~with the axioms for comparatives.
}
\label{flow}
\end{figure*}
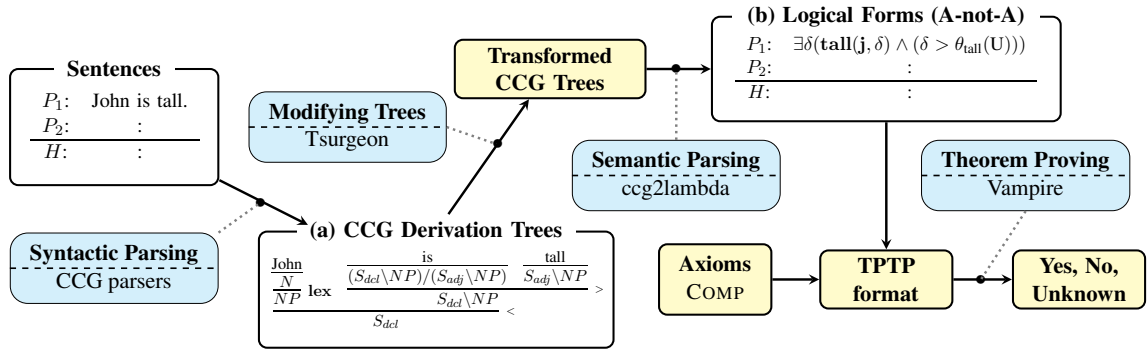

In another approach, unsupervised NLI systems based on various logics have been studied~\citep{bos2008wide,maccartney2008modeling,mineshima2015higher,abzianidze2016natural}.
However, the accuracies of these systems on comparative constructions are relatively low (see Section \ref{sec:experiment}).
Although there
have been
detailed discussions in formal semantics
taking
into account the complexity associated with adjectives and comparative expressions~\citep{cresswell1976semantics,kennedy97,heim2000degree,lassiter2017graded}, such theories have not yet been implemented in NLI systems.
Also, some logic-based NLI systems
handle comparatives~\citep{chatzikyriakidis-bernardy-2019-wide, haruta2019:paclic},
but these systems do not implement a parser and/or a prover.

The goal of this study is to fill this gap by implementing a formal compositional semantics based on the so-called A-not-A analysis~\citep{seuren1973comparative,klein1980semantics,klein1982interpretation,schwarzschild2008semantics},
which maps various comparative constructions in English to logical forms (LFs)
via CCG~\citep{Steedman2000} derivation trees.
Based on this, we present 
an inference system that
computes complex logical inference over  
comparatives, generalized quantifiers, and numerals.\footnote{GitHub repository with code and data: \url{https://github.com/izumi-h/ccgcomp}}
For evaluation,
we use the FraCaS test set~\citep{cooper1994fracas}, which
contains various linguistically challenging inferences,
and the Monotonicity Entailment Dataset (MED)~\citep{yanaka-etal-2019-neural},
which contains inferences with generalized quantifiers.
We also construct a new test set, the Comparative and Adjective Dataset (CAD), 
which extends FraCaS and collects both single-premise and multi-premise inferences with comparatives.
The experiments show that
our system outperforms previous logic-based systems as well as recent DL models.

\begin{table*}[t]
  \centering
  \scalebox{0.84}{$
  {\renewcommand\arraystretch{1.1}
  \begin{tabular}{c|l|l|l}\bhline{1.3pt}
    \textbf{Pattern} &
    \textbf{Example} &
    \textbf{Type} &
    \textbf{LF}\\
    \hline\hline
    \multirow{3}{*}{(i)} & 1. John is tall. & Positive & $\exists \delta(\funtall{\LF{john}}{\delta} \wedge (\delta >  \funtheta{tall}{U}))$\\
     & 2. John is taller than Bob. & Increasing & \funtaller{\LF{john}}{\LF{bob}}\\
     & 3. Ann has more children than Bob. & Numerical & $\exists \delta(\exists x(\fun{child}{x} \wedge \funt{have}{\LF{ann}}{x} \wedge \funt{many}{x}{\delta})$\\
     & & & $\wedge \neg\exists x(\fun{child}{x} \wedge \funt{have}{\LF{bob}}{x} \wedge \funt{many}{x}{\delta}))$\\
    \hline
    \multirow{3}{*}{(ii)} & 1. John is as tall as Bob. & Equatives & \funasadj{tall}{\LF{bob}}{\LF{john}}\\
     & 2. Mary is 2 inches taller than Harry. & Differential & $\forall \delta(\funtall{\LF{harry}}{\delta - 2^{\prime\prime}} \rightarrow \funtall{\LF{mary}}{\delta})$\\
     & 3. John ate 3 more cookies than Bob. & Measure & $\forall \delta(\exists x(\fun{cookie}{x} \wedge \funt{eat}{\LF{bob}}{x} \wedge \funt{many}{x}{\delta - 3})$\\
     & & & $\rightarrow \exists x(\fun{cookie}{x} \wedge \funt{eat}{\LF{john}}{x} \wedge \funt{many}{x}{\delta}))$\\
    \bhline{1.3pt}
    \end{tabular}
    }
    $}
    \caption{Semantic representation of comparative constructions based on A-not-A analysis}
    \label{tab:comparatives}
\end{table*}

\section{System overview}

Figure \ref{flow} shows the pipeline of the proposed system.
First, the input sentences are a set of premises $P_1, \ldots, P_n$ and a hypothesis $H$.
Next, the
CCG derivation trees are obtained
using CCG parsers.
Derivation trees are modified to derive appropriate LFs based on A-not-A analysis.
We use
the
semantic parsing system ccg2lambda~\citep{martinez-gomez-etal-2016-ccg2lambda} based on $\lambda$-calculus 
to obtain LFs,
which are then converted to 
the Typed First-order Form (TFF) of the Thousands of Problems for Theorem Provers (TPTP) format~\citep{Sut17},
that is, a formal expression in first-order logic with equality and arithmetic operations.
Finally, together with the axiom system \comp~\citep{haruta2019:paclic} for 
comparatives and numerical expressions,
a theorem prover checks whether
$P_1 \wedge \cdots \wedge P_n \to H$ holds or not.
The system output is \LabelYes~(entailment),
\LabelNo~(contradiction), or \LabelUnk~(neutral).

\subsection{Degree semantics: A-not-A analysis}
\label{sec:anota}

In formal semantics, comparative and other gradable expressions are usually analyzed using the notion of \textit{degree}~\citep{cresswell1976semantics}.
\begin{exe}
    \ex \label{basic}
    \begin{xlist}
        \ex Ann is \textit{taller} than Bob.
        \ex John is \textit{5 feet tall}.
        \ex John is \textit{tall}.
    \end{xlist}
\end{exe}

\noindent
For example, the sentence (\ref{basic}a), in which the comparative form \textit{taller} of the gradable adjective \textit{tall} is used, compares the degree of height between two persons.
(\ref{basic}b) is an expression that includes a specific height, which is the numerical expression \textit{5 feet}.
(\ref{basic}c) is a sentence using the positive form of the adjective, which 
can be regarded as representing a comparison with some implicit standard value.
In degree-based semantics, such gradable adjectives are treated as two-place predicates
that have
entity and degree~\citep{cresswell1976semantics}.
For instance, (\ref{basic}b) is analyzed as \funtall{\LF{john}}{\LF{5~feet}}, where $\funtall{x}{\delta}$ is read as
``$x$ is \textit{at least} as tall as degree $\delta$''~\citep{klein91}.

We use A-not-A analysis of comparatives, which analyzes (\ref{anota}a) as (\ref{anota}b).

\begin{exe}
    \ex \label{anota}
    \begin{xlist}
        \ex Ann is taller than Bob is. 
        \ex $\funtaller{\LF{ann}}{\LF{bob}}$ 
    \end{xlist}
\end{exe}

\begin{center}
\scalebox{0.74}{
    \begin{tikzpicture}
        \node(nA) at (0,0) {Ann};
        \node(nB) at (0,-0.7) {Bob};
	    \node[fill=green!25, rectangle, draw=black, line width=1pt, text width=5.5cm,text centered,minimum height=1em, right=0.5em of nA] (A) {};
	    \node[fill=green!25, rectangle, draw=black, line width=1pt, text width=4cm,text centered,minimum height=1em, right=0.5em of nB] (B) {};
	    \coordinate (O) at (0.45,-1.3) node at (O) [left] {0};
	    \coordinate (P) at (6.7,-1.3);
	    \draw [->, line width=1pt] (O)--(P) node at (P) [right] {$\delta$};
	    \draw [densely dashed, line width=1pt] (6.43,0)--(6.43,-1.3) node at (6.24,-1.3) [below] {\large $\delta_1$};
	    \draw [densely dashed, line width=1pt] (4.92,-0.7)--(4.92,-1.3) node at (4.73,-1.3) [below] {\large $\delta_2$};
	    \draw [densely dashed, line width=1pt] (5.69,0.17)--(5.69,-1.3) node at (5.5,-1.3) [below] {\large $\delta$};
    \end{tikzpicture}
}
\end{center}
According to this analysis, (\ref{anota}a) is interpreted as saying that
there exists a degree $\delta$ of height 
that Ann satisfies, but Bob does not.
As shown in the figure in (\ref{anota}),
this guarantees that Ann's height is greater than Bob's height.
A-not-A analysis makes it possible to
derive entailment relations between various comparative
constructions in a simple way 
using first-order logic theorem provers.

Table \ref{tab:comparatives} shows LFs
for some example sentences
using
A-not-A analysis.\footnote{
For the positive form, the comparison class~\citep{klein1982interpretation} is relevant
to determining the standard of degree (e.g., tallness). We use a default comparison class such as $\theta_{\text{tall}}$ in our implementation
and leave the determination of comparison classes and relevant standards~\citep[cf.][]{pezzelle-fernandez-2019-red} to future work. 
}\
Here, LFs can be divided into two patterns.
The examples in (i) in Figure \ref{tab:comparatives}
belong
to the first type,
where the degree of an individual
\textbf{exceeds} a certain degree.
For example, the sentence (i-2) means that the height of John is greater than the height of Bob.
The sentence (i-3) means that 
the number of Ann's children
exceeds the number of Bob's children.
Under our analysis, this type of sentence
is mapped to formulas of the form $\exists \delta(\cdots \wedge \cdots)$.

The second type includes the examples in (ii),
which say that the degree of an individual is \textbf{greater than or equal to} a certain degree.
For example, (ii-1) means that John's height is greater than or equal to Bob's height~\citep{klein1982interpretation}.
The sentence (ii-3) means that 
the number of cookies John ate is 3 or more greater than the number of cookies that Bob ate;
in other words,
if Bob ate $n$ cookies,
then John ate at least $n + 3$ cookies.
Sentences of type (ii) are mapped to
formulas of the form $\forall \delta(\cdots \to \cdots)$, as in Table \ref{tab:comparatives}.

\begin{figure*}[t]
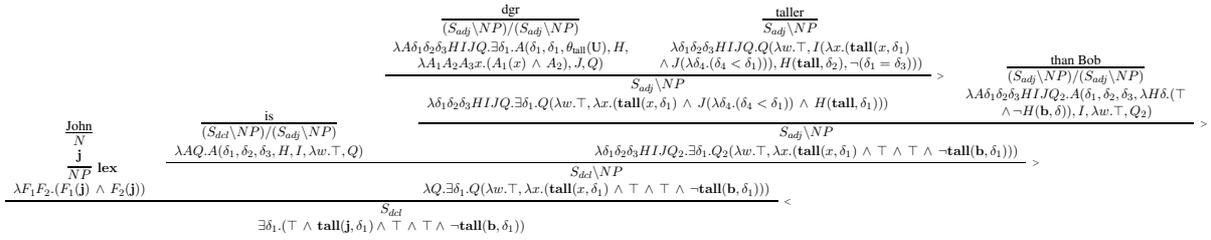

\scalebox{0.47}{
\deduce{\SR{\exists \delta_1.(\top \,\wedge\, \LF{tall}(\LF{j},\delta_1) \wedge\,\top \,\wedge\,\top \wedge\, \neg \LF{tall}(\LF{b},\delta_1))}}{
\infer[\BA]{S_{dcl}}{
\deduce{\SR{\lambda F_1 F_2.(F_1(\LF{j}) \,\wedge\, F_2(\LF{j}))}}{
\infer[\LF{lex}]{NP}{
\deduce{\SR{\LF{j}}}{
\infer{N}{\mbox{John}}}}} & 
\deduce{\SR{\lambda Q.\exists \delta_1.Q(\lambda w.\top,\lambda x.(\LF{tall}(x,\delta_1) \,\wedge\,\top \,\wedge\,\top \,\wedge\, \neg \LF{tall}(\LF{b},\delta_1)))}}{
\infer[\FA]{S_{dcl}\backslash NP}{
\deduce{\SR{\lambda A Q.A(\delta_1,\delta_2,\delta_3,H,I,\lambda w.\top,Q)}}{
\infer{(S_{dcl}\backslash NP)/(S_{adj}\backslash NP)}{\mbox{is}}} &
\deduce{\SR{\lambda \delta_1 \delta_2 \delta_3 H I J Q_2.\exists \delta_1.Q_2(\lambda w.\top,\lambda x.(\LF{tall}(x,\delta_1) \,\wedge\,\top \,\wedge\,\top\,\wedge\, \neg \LF{tall}(\LF{b},\delta_1)))}}{
\infer[\FA]{S_{adj}\backslash NP}{
\deduce{\SR{\lambda \delta_1 \delta_2 \delta_3 H I J Q.\exists \delta_1.Q(\lambda w.\top,\lambda x.(\LF{tall}(x,\delta_1)\,\wedge\, J(\lambda \delta_4.(\delta_4<\delta_1))\,\wedge\, H(\LF{tall},\delta_1)))}}{
\infer[\FA]{S_{adj}\backslash NP}{
\deduce{\SR{\lambda A \delta_1 \delta_2 \delta_3 H I J Q.\exists \delta_1.A(\delta_1,\delta_1,\funtheta{tall}{U},H,$\\$\lambda A_1 A_2 A_3 x.(A_1(x) \,\wedge\, A_2),J,Q)}}{
\infer{(S_{adj}\backslash NP)/(S_{adj}\backslash NP)}{\mbox{dgr}}} & 
\deduce{\SR{\lambda \delta_1 \delta_2 \delta_3 H I J Q.Q(\lambda w.\top,I(\lambda x.(\LF{tall}(x,\delta_1) $\\$\,\wedge\, J(\lambda \delta_4.(\delta_4<\delta_1))),H(\LF{tall},\delta_2),\neg (\delta_1 = \delta_3)))}}{
\infer{S_{adj}\backslash NP}{\mbox{taller}}}}}
&
\deduce{\SR{\lambda A \delta_1 \delta_2 \delta_3 H I J Q_2.A(\delta_1,\delta_2,\delta_3,\lambda H \delta.(\top$\\$\,\wedge\, \neg H(\LF{b},\delta)),I,\lambda w.\top,Q_2)}}{
\infer{(S_{adj}\backslash NP)/(S_{adj}\backslash NP)}{\mbox{than Bob}}}}}}}}}
}
\caption{Derivation tree of \textit{John is }(\textit{dgr})\textit{ taller than Bob.}}
\label{fig:dgr}
\end{figure*}

\begin{table*}[t]
  \centering
  \scalebox{0.77}{$
  {\renewcommand\arraystretch{1.1}
  \begin{tabular}{l|l}\bhline{1.3pt}
    \textbf{Example} &
    \textbf{LF}\\
    \hline\hline
    Mary has \textit{many} dogs. & $\exists x(\funt{have}{\LF{mary}}{x} \wedge \fun{dog}{x} \wedge \funt{many}{x}{\theta_{\text{many}}(x)})$\\
    Ann read \textit{two} books. & $\exists x(\funt{read}{\LF{ann}}{x} \wedge \fun{book}{x} \wedge \funt{many}{x}{2})$\\
    \textit{Most} apples are red. & 
    $\exists \delta(\exists x(\fun{apple}{x} \wedge \fun{red}{x} \wedge \funt{many}{x}{\delta}) \wedge \neg\exists x(\fun{apple}{x} \wedge \neg\fun{red}{x} \wedge \funt{many}{x}{\delta}))$
    \\
    \textit{No more than five} boys ran. & 
    $\neg\exists x \exists \delta(\fun{boy}{x} \wedge \funt{many}{x}{\delta} \wedge (5 < \delta) \wedge \fun{run}{x})$
    \\
    \bhline{1.3pt}
    \end{tabular}
    }
    $}
    \caption{LFs of generalized quantifiers based on our degree semantics}
    \label{tab:gq}
\end{table*}

\subsection{Compositional semantics in CCG}
\label{sec:ccg}

In CCG, the mapping from syntax to semantics is defined by assigning syntactic categories to words~\citep{Steedman2000}; the LF of a sentence is then compositionally derived using $\lambda$-calculus.
However, there is a gap between the syntactic structures assumed in formal semantics
and the output derivation trees of
existing CCG parsers, i.e., statistical parsers
trained on CCGBank~\citep{HockenmaierSteedman}.
For this reason, we modify
the derivation trees provided by CCG parsers in post-processing.
There are several types of modifications.

\paragraph{Syntactic features}
The first modification is
to add
syntactic features to CCG categories.
For example, in the default CCG trees,
a nominal adjective (\textit{a \underline{tall} boy}) has
the category $N/N$,
while a predicate
adjective (\textit{John is \underline{tall}})
has the category $S_{adj}\backslash NP$.
To provide a uniform degree semantics to both constructions, we rewrite $N/N$ as $N_{adj}/N$ for the category of nominal adjectives.

\paragraph{Multiword expressions}
Compound expressions for 
comparatives and quantifiers are combined as one word, such as \textit{a few, a lot of}, and \textit{at most}.

\paragraph{Empty categories}
We insert an empty category to systematically derive the LFs of the two patterns described in Table \ref{tab:comparatives}.
The distinction between
patterns (i) and (ii) can be controlled by an expression appearing in the adjunct position of an adjective phrase,
for example, a degree modifier such as \textit{very} or a numerical expression such as \textit{2 cm}.
When such an adjunct expression does not appear,
we insert an empty category \textit{dgr} into the adjunct position,
which is used to derive the desired LF compositionally.
Figure \ref{fig:dgr} shows an example
of a modified derivation tree containing an empty element \textit{dgr} for increasing comparatives.
Similarly, we use two other types of empty categories for equatives (e.g., \textit{as tall as})
and the positive form.

\subsection{Generalized quantifiers}
\label{sec:gq}
The analysis of comparatives by the degree-based semantics described above can naturally be extended to generalized quantifiers.
In the
traditional
analysis~\citep{Barwise1981-BARGQA},
generalized quantifiers such as
\textit{many, few, more than}, and \textit{most} are analyzed as denoting a relation between sets.
Alternatively, an analysis based on degree semantics has been
developed, which represents
expressions such as \textit{many} and \textit{few}
as adjectives~\citep{partee1988many,rett2018semantics}
and \textit{most}
as the superlative form of \textit{many}~\citep{hackl2000comparative,szabolcsi2010quantification}.
We recast this alternative analysis in our degree-based semantics. Table \ref{tab:gq}
shows the LFs for some examples.
We use the binary predicate \funt{many}{x}{n},
which reads ``$x$ is composed of (at least) $n$ entities''.
\textit{Most A are B} is analyzed as meaning ``More than half of \textit{A} is \textit{B}'', following the
standard
truth-condition~\citep{hackl2000comparative}.

\section{Experiments}
\label{sec:experiment}

\subsection{Experimental settings}

For CCG parsing, we use two CCG parsers,
namely,
C\&C~\citep{Clark2007}
and depccg~\citep{yoshikawa-etal-2017-ccg},
to mitigate parsing errors.
If two parsers output a different answer,
we choose the system answer in the following way: 
if one answer is \LabelYes~(resp.~\LabelNo) and the other is \LabelUnk, the system answer is \LabelYes~(resp.~\LabelNo);
if one answer is \LabelYes\ and the other is \LabelNo,
then the system answer is \LabelUnk.
For POS tagging, we use
the C\&C POS tagger for C\&C
and spaCy\footnote{
\url{https://github.com/explosion/spaCy}
} for depccg.

To implement compositional semantics, we use ccg2lambda\footnote{\url{https://github.com/mynlp/ccg2lambda}}.
We extend the semantic templates
proposed in \citet{mineshima2015higher}
to handle linguistic phenomena based on degree-based semantics.
The total number of lexical entries assigned to CCG categories is 106, and
the number of entries directly assigned to particular words 
(e.g., \textit{than} and \textit{as} for comparatives and items for quantifiers) is 214.
For tree transformation, we use
Tsurgeon~\citep{levy-andrew-2006-tregex}.
We use 74 entries (rewriting clauses) in the Tsurgeon script.
For theorem proving, we use
Vampire\footnote{\url{ https://github.com/vprover/vampire}}, which accepts TFF forms with arithmetic operations.

For evaluation, we use three datasets.
First, FraCaS~\citep{cooper1994fracas} is a dataset
comprising
nine sections, each of which
contains
semantically challenging inferences
related to
various linguistic phenomena.
In this study, we use three sections: Generalized Quantifiers (\textsc{GQ}; 73 problems), Adjectives (\textsc{Adj}; 22 problems), and Comparatives (\textsc{Com}; 31 problems).
The distribution of gold answer labels for the three sections is (\LabelYes/\LabelNo/\LabelUnk) = (36/5/32), (9/6/7), (19/9/3), respectively.

Second, MED\footnote{\url{https://github.com/verypluming/MED}} is a dataset that contains inferences with quantifiers (so-called monotonicity inferences).
We use a subset (498 problems) of MED that does not require world knowledge and commonsense reasoning;
these problems were collected from various linguistics papers.
The distribution of the gold answer is
(\LabelYes/\LabelUnk) = (215/283).

Because there are only 31 problems for comparatives in FraCaS,
we created
the CAD test set
consisting of 105 problems,
which focuses on comparatives and numerical constructions not covered by FraCaS.
We collected a set of inferences (9 problems) from a linguistics paper~\citep{klein1982interpretation}
and created more problems by
adding negation,
using
degree modifiers (e.g., \textit{very}), changing numerical expressions,
replacing positive and negative adjectives (e.g., ~\textit{large} to \textit{small}), and swapping the premise and hypothesis of an inference.
Of the 105 problems 50 are single-premise problems, and 55 are multi-premise problems.
The distribution of gold answer labels is (\LabelYes/\LabelNo/\LabelUnk) = (50/17/38).
All of the gold labels were checked by an expert in linguistics.
Table \ref{tab:fracas} shows some example problems.
\begin{table}[t]
\small
 \centering
\scalebox{0.9}{$\displaystyle
    {\renewcommand\arraystretch{1.3}
  	\begin{tabular}{l|l}\bhline{1.3pt}
    	\multicolumn{2}{l}{FraCaS-235 (\textsc{comparatives}) \ Gold answer: Yes} \\
    	\hline
    	\textbf{Premise 1} & ITEL won more orders than APCOM.\\
    	\textbf{Premise 2} & APCOM won ten orders.\\
    	\textbf{Hypothesis} & ITEL won at least eleven orders.\\
    	\hline\hline
    	\multicolumn{2}{l}{MED-1085 \ Gold answer: Unknown} \\
    	\hline
    	\textbf{Premise 1} & No more than fifty campers have caught a cold.\\
    	\textbf{Hypothesis} & No more than fifty campers have had a sunburn \\
    	& or caught a cold.\\
    	\hline\hline
        \multicolumn{2}{l}{CAD-011 (\textsc{comparatives}) \ Gold answer: Yes} \\
    	\hline
    	\textbf{Premise 1} & Alex is not as tall as Chris is.\\
    	\textbf{Hypothesis} & Chris is taller than Alex is.\\
        \hline\hline
        \multicolumn{2}{l}{CAD-034 (\textsc{adjectives}) \ Gold answer: Yes} \\
    	\hline
    	\textbf{Premise 1} & Bob is 4 feet tall.\\
    	\textbf{Premise 2} & John is taller than Bob.\\
    	\textbf{Hypothesis} & John is more than 4 feet tall.\\
    	\bhline{1.3pt}
  	\end{tabular}
  	}
  	$}
  	\caption{Examples of entailment problems from the FraCaS, MED, and CAD test sets}
  	\label{tab:fracas}
\end{table}

\begin{table}[t]
    \centering
    \scalebox{0.85}{$\displaystyle
	    \begin{tabular}{c|l|ccc}
	        \bhline{1.3pt}
	        \multicolumn{5}{l}{FraCaS}\\
	        \hline\hline
	        \multicolumn{2}{l|}{Section} & \textsc{GQ} & \textsc{Adj} & \textsc{Com}\\
	        \hline
	        \multicolumn{2}{l|}{\#All} & 73 & 22 & 31\\
	        \multicolumn{2}{l|}{\#Single} & 44 & 15 & 16\\
	        \hline
	        \multicolumn{2}{l|}{Majority} & .48 & .39 & .61\\
	        \hline
	        \multirow{4}{*}{Logic} & \texttt{MN} & .77 & .68 & .48\\
	        & \texttt{LP} & .93 & .73 & -\\
	        & \texttt{NL} & ~~.98* & ~~.80* & ~~.81*\\
	        & Ours & .92 & .86 & .77\\
	        & ~+rule & \textbf{.95} & \textbf{.95} & \textbf{.84}\\
	        \hline
	        \multirow{3}{*}{DL} & \texttt{LSTM} & ~~.64* & ~~.47* & ~~.56*\\
	        & \texttt{DA} & .59 & .45 & .61\\
	        & \texttt{BERT} & .64 & .45 & .58\\
            \bhline{1.3pt}
  	    \end{tabular}
    $}
    \caption{Accuracy on the FraCaS test suite: `\#All' shows the number of all problems and `\#Single' the number of single-premise problems.
    }
  	\label{tab:fresult}
\end{table} 

\subsection{Results and discussion}

\paragraph{FraCaS test suite}\ \
Table~\ref{tab:fresult} shows the experimental results on FraCaS.
\textit{Majority} is the accuracy of the majority baseline and \textit{Ours} the accuracy of our system.
Some errors were caused by failing to assign
correct POS tags and lemmas to comparatives;
for example, \textit{cleverer} is wrongly assigned \textsl{NN} rather than \textsl{JJR} (FraCaS-217). To estimate the upper bound of the accuracy of our system by reducing error propagation, we added hand-coded rules to assign correct POS tags and lemmas (14 words).
We also added two rules to join
multiword expressions to derive
correct logical forms (\textit{law lecturer} and \textit{legal authority} for FraCaS-214, 215).
In Table \ref{tab:fresult}, \textit{+rule} shows
the improvement in accuracy
realized
by adding these rules.

We compare our system with
previous logic-based NLI systems as well as
three popular DL models.
For logic-based systems,
we use \texttt{MN}~\citep{mineshima2015higher} and \texttt{LP}~\citep{abzianidze2016natural} based on CCG parsers and theorem proving and \texttt{NL}~\citep{maccartney2008modeling} based on Natural Logic. \texttt{NL} is evaluated on single-premise problems only (indicated by *).
Our system accepts both single-premise and multiple-premise problems and 
outperforms the previous logic-based systems
on the adjectives and comparatives sections.
Our system solves complex reasoning problems with multiple premises involving comparatives and numerical expressions, such as FraCaS-235 in Table~\ref {tab:fracas},
for which the previous systems
were unable
to give a correct answer.

For DL models, \texttt{LSTM} is the performance of a
long short-term memory
model trained on SNLI, which is reported in~\citet{bowman2016modeling} (only evaluated on single-premise problems).
We also tested
the Decomposable Attention (\texttt{DA}) model ~\citep{parikh-etal-2016-decomposable}, a
simple attention-based model
trained on SNLI.
We used the implementation provided in AllenNLP~\citep{gardner-etal-2018-allennlp}.
Finally, \texttt{BERT}
is the performance of a BERT model~\citep{devlin-etal-2019-bert}.
We used the 
\texttt{bert-base-cased}
model fine-tuned with
MultiNLI.
We used the code available at the original GitHub repository.\footnote{\url{https://github.com/google-research/bert}}
Our system outperforms the three DL models by large margins.\footnote{For \texttt{DA} and \texttt{BERT},
we evaluated multiple-premise problems
by two methods: simply concatenating two or more premises (e.g., ``\textit{S$_1$. S$_2$.}'') and by inserting \textit{and} and commas between sentences (e.g., ``\textit{S}$_1$ \textit{and} \textit{S}$_2$.'').
Comparing the two methods, we used the better accuracy for each problem in MED and CAD in Table \ref{tab:fresult} and \ref{tab:med}.
}

\begin{table}[t]
    \centering
    \scalebox{0.93}{$\displaystyle
        \begin{tabular}{l|c}
	        \bhline{1.3pt}
	        \multicolumn{2}{l}{MED}\\
	        \hline\hline
	        \#All & 498\\
	        \hline
	        Majority & .60\\
	        \hline
	        \textbf{\texttt{BERT+}} & .54\\
	        \textbf{\texttt{BERT}} & .56\\
	        Ours & \textbf{.84}\\
            \bhline{1.3pt}
        \end{tabular}
    $}
    \hspace{2em}
    \centering
    \scalebox{0.9}{$\displaystyle
	    \begin{tabular}{l|c}
	        \bhline{1.3pt}
	        \multicolumn{2}{l}{CAD}\\
	        \hline\hline
	        \#All & 105\\
	        \hline
	        Majority & .48\\
	        \hline
	        \texttt{DA} & .51\\
	        \texttt{BERT} & .55\\
	        Ours & \textbf{.77}\\
            \bhline{1.3pt}
        \end{tabular}
    $}
    \caption{Accuracy on the MED and CAD datasets}
    \label{tab:med}
\end{table}  
\paragraph{MED and CAD datasets}\ \
Table \ref{tab:med} shows the results on MED and CAD.
For MED, we compared our system with
a BERT model fine-tuned with MultiNLI (\textbf{\texttt{BERT}})
and a BERT model with data augmentation
(approximately 36K) in addition to MultiNLI (\textbf{\texttt{BERT}+}), both being tested in \citet{yanaka-etal-2019-neural}.
For CAD, we evaluated \texttt{DA} and \texttt{BERT}.
The results show that our system achieved high accuracy on the logical inferences with adjectives, comparatives, and generalized quantifiers.

Table \ref{tab:nnmodel} shows examples that were solved by our system but not by \texttt{DA} and \texttt{BERT}.
The DL models were particularly difficult to handle inferences related to antonyms (e.g., FraCaS-209) and numerical expressions (e.g., CAD-001).
Indeed, the results on the DL models were
predictable because
these models were trained on datasets (SNLI and MultiNLI) 
that do not target the logical and numerical inferences we are concerned with in this study.
However, it is fair to say that it is very challenging to generate effective training data 
to handle
various complex inferences with comparatives, numerals, and generalized quantifiers.

\begin{table}[t]
\small
\smallskip
 \centering
 \scalebox{0.93}{$\displaystyle
    {\renewcommand\arraystretch{1.3}
  	\begin{tabular}{l|l}\bhline{1.3pt}
    	\multicolumn{2}{l}{FraCaS-209 (\textsc{adjectives}) \ Gold answer: No} \\
    	\hline
    	\textbf{Premise 1} & Mickey is a small animal.\\
    	\textbf{Premise 2} & Dumbo is a large animal.\\
    	\textbf{Hypothesis} & Mickey is larger than Dumbo.\\
    	\hline\hline
    	\multicolumn{2}{l}{MED-1021 \ Gold answer: Unknown} \\
    	\hline
    	\textbf{Premise 1} & More than five campers have had a sunburn \\
    	& or caught a cold. \\
    	\textbf{Hypothesis} & More than five campers have caught a cold. \\
        \hline\hline
        \multicolumn{2}{l}{CAD-001 \ Gold answer: Yes} \\
    	\hline
    	\textbf{Premise 1} & John is 5 cm taller than Bob.\\
    	\textbf{Premise 2} & Bob is 170 cm tall.\\
    	\textbf{Hypothesis} & John is 175 cm tall.\\
    	\hline\hline
        \multicolumn{2}{l}{CAD-103 \ Gold answer: Unknown} \\
    	\hline
    	\textbf{Premise 1} & Bob is not tall.\\
    	\textbf{Premise 2} & John is not tall. \\
    	\textbf{Hypothesis} & John is taller than Bob.\\
    	\bhline{1.3pt}
  	\end{tabular}
  	}
  	$}
  	\caption{Examples of problems solved by our system but not by the DL models.
  	The answers of the DL models are:
        \textit{yes} (\texttt{DA} and \texttt{BERT}) for FraCaS-209;
        \textit{yes} (\textbf{\texttt{BERT}} and \textbf{\texttt{BERT+}}) for MED-1021;
        \textit{no} (\texttt{DA} and \texttt{BERT}) for CAD-001;
        \textit{yes} (\texttt{DA}) and \textit{no} (\texttt{BERT}) for CAD-103.
}
  	\label{tab:nnmodel}
\end{table}

There
were some
problems that our system could not solve. 
For FraCaS, the accuracy for the comparative section (\textsc{Com}) was
relatively low ($.84$).
This is because this section contains linguistically challenging phenomena 
such as clausal comparatives (FraCaS-239, 240, 241)
and attributive comparatives (FraCaS-244, 245).
For MED, the present system does not handle
downward monotonic quantifiers (e.g., \textit{less than}), non-monotonic quantifiers (e.g., \textit{exactly}),
and negative polarity items (e.g., \textit{any}).
Furthermore,
the system needs to be extended
to deal with linguistic phenomena such as comparative subdeletion and quantified comparatives that appear in CAD.
To
address
these problems, 
further improvement of the CCG parsers will be needed.

\section{Conclusion}

In this study, we presented an end-to-end logic-based inference system for
handling complex inferences with
comparatives, quantifiers, and numerals.
The entire system is transparently composed of several modules and can solve complex inferences for the right reason.
In future work, we will extend our analysis to cover the more complex constructions mentioned in Section \ref{sec:experiment}.
We are also considering combining
our system
with an abduction mechanism
that uses large knowledge bases~\citep{yoshikawa2019combining} for handling commonsense reasoning with external knowledge.

\paragraph{Acknowledgments}
We are grateful to Hitomi Yanaka 
for sharing the detailed results on the MED dataset
and Masashi Yoshikawa for continuous support.
We also thank the three anonymous reviewers for their helpful comments and feedback.
This work was supported by JSPS KAKENHI Grant Number JP18H03284.

\bibliography{refs}
\bibliographystyle{acl}

\noautomath
\medskip

\end{document}